\documentclass[runningheads]{llncs}
\usepackage{graphicx}
\usepackage{amsmath}

\begin{document}
\title{An Approach for Improving Automatic Mouth Emotion Recognition}
 \author{Giulio Biondi\inst{1,2}$^{ORCID: 0000-0002-1854-2196}$ \and Valentina Franzoni\inst{2,3,4}$^{ORCID: 0000-0002-2972-7188}$ \and Osvaldo Gervasi\inst{2}$^{ORCID: 0000-0003-4327-520X}$ \and Damiano Perri\inst{2}$^{ORCID: 0000-0001-6815-6659}$
 }

\institute{
University of Florence, Dept. of Mathematics and Computer Science,\\ Florence, Italy\\
\and
University of Perugia, Dept. of Mathematics and Computer Science,
Perugia, Italy\\
\and
Sapienza University of Rome, Department of Computer, Control, and Management Engineering \lq\lq Antonio Ruberti\rq\rq, Rome, Italy
\and
\textit{corresponding author}
}
\titlerunning{An Approach for Improving Automatic Mouth
Emotion Recognition} 
\authorrunning{G. Biondi, V. Franzoni, O. Gervasi, D. Perri} 

\maketitle
\begin{abstract}
The study proposes and tests a technique for automated emotion recognition through mouth detection via Convolutional Neural Networks (CNN), meant to be applied for supporting people with health disorders with communication skills issues (e.g. muscle wasting, stroke, autism, or, more simply, pain) in order to recognize emotions and generate real-time feedback, or data feeding supporting systems. The software system starts the computation identifying if a face is present on the acquired image, then it looks for the mouth location and extracts the corresponding features. 
Both tasks are carried out using Haar Feature-based Classifiers, which guarantee fast execution and promising performance. 
If our previous works focused on visual micro-expressions for personalized training on a single user, this strategy aims to train the system also on generalized faces data sets.
\end{abstract}

\section{Introduction}
In this work, we present a system for mouth-based visual emotion recognition. Our purpose is to lay the basis for a health-care system for people who suffer from severe disease, e.g., strokes, or conditions such as autism, who may benefit from automated support of emotion recognition. Such systems can detect basic emotions from smartphone or computer camera devices, to produce feedback, either text, audio or visual for other humans, or a digital output to support other connected services. Connecting such an architecture to appropriate services could help users to convey their or others' emotions more effectively, providing augmented emotional stimuli, e.g., in case of users affected from a pathology which involve social relationship abilities, or when users experiment difficulties in recognizing emotions expressed by others. The system could also call a human assistant, e.g., for hospitalized patients feeling intense pain. In this paper, we focus on the mouth expression in correctly determining the emotion expressed by a subject.
A crucial step is the selection of a reference model which classifies emotions, e.g., Ekman,\cite{Ekman} Plutchik, and Lovheim \cite{Poggioni2013}. We selected a basic subset of the Ekmann emotions: \textit{Joy} and \textit{Disgust}, together with the \textit{Neutral} condition (i.e., no emotion expressed). Joy is among the simplest emotions to recognize through face expression, thus an ideal candidate for results comparisons concerning the state-of-the-art. Disgust, instead, is present in much fewer instances in available data sets, because it is more difficult to stimulate, and it is a less ideal but more interesting example of computation. We include the neutral state as a control state for recognition results on both emotions.
\section{Problem description and proposed solution}
Our study exploits the high precision of CNN processing to process mouth images to recognize emotional states. On one hand, we expect the system's capability to exploit best on the single user with personalized training; on the other hand, in this work we also test the technique on generic faces data sets, in order to find solutions to the following research questions:\\
\textit{- With which precision it is possible to recognize facial emotions solely from the mouth?}\\
\textit{- Is the proposed technique capable of recognizing emotions if trained on a generalized set of facial images?}\\
In a user-centered implementation, the user trains the network on her/his facial expressions and the software supports personalized emotional feedback for each particular user: personal traits, such as scars or flaws, or individual variations in emotional feeling and expression, help the training to precise recognition. Then, we train the software also to recognize different users.
In order to obtain optimized results, the ambient light setting needs a proper setup:
\begin{itemize}
\item \textbf{Robustness}: The algorithm must be able to operate even in the presence of low-quality data (e.g., low resolution, bad light conditions);

\item \textbf{Scalability}: The user position should not be necessarily fixed in front of the camera, in order to avoid constraining the person. Therefore, the software should be able to recognize the user despite her/his position.

\item \textbf{Luminosity}: an important problem is precisely that of the variation of light. In computer vision (\cite{Cootes1995}), the variation of the lens involves an alteration of the information (\cite{Franzoni2018}. No complete control of the detected information is achieved: the system will be able to withstand variations in brightness without compromising the original information. 
\end{itemize}

The proposed solution has been implemented in C++ and OpenCV graphics libraries; hence, it is compatible with all operating systems, with high reliability and constant support from the community.
\section{State of the Art}
\subsection*{Deep Learning and Image Classification}
The recent scientific focus on Deep Learning towards the end of the XX century has contributed to the rebirth of significant interest in neural networks. The real impact of Deep Learning began in the context of speech recognition around the year 2010, when two Microsoft Research employees, Lil Deng and Geoenix Hinton, realized that using large amounts of data for training a deep neural network resulted in lowering error rates far below the state of the art \cite{LeCun2015}.
Discoveries in the field of hardware have certainly contributed to the rise of interest in Deep Learning. In particular, the ever-powerful GPUs seem to be able to perform the countless mathematical calculations of matrices and vectors in Deep Learning \cite{VellaGPU,VellaNGT12,VellaFGCS}. Actual GPUs allow reducing workout times from the weeks to a day.
Recently, deep learning has been used for several types of research aiming at the classification of images and learning, trying to solve the limitations of machine learning, which reside in overfitting and domain dependence, with image adaptation, kernel randomization \cite{Novi} and transfer learning \cite{Transfer}.
Commitment has been dedicated by researchers to exploit domain dependence as a feature, where personalized classification can quickly exploit a particular user or entity, especially for smart-home systems \cite{smarthome} and microblog sentiment tagging \cite{personalizedmicroblog}.
Alternative approaches consider evolutionary algorithms,\cite{semanticcontext}\cite{reactiveEnvironments}\cite{pheromone} random walks on semantic networks of images \cite{imagesimilarity}\cite{semanticgraphs}\cite{gridservices} and max-product neural networks.

\subsection*{History and Description of Neural Networks}
\begin{figure}
\centering
\includegraphics[scale=0.6]{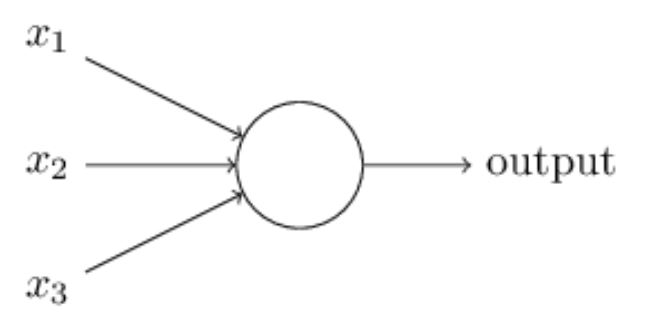}
\caption{A simple example of a 3-input perceptron}
\label{fig:NNoutput}
\end{figure}

Convolutional Neural Networks are among the most used methods for affective image classification \cite{emocnn}\cite{Stiefelhagen1995} thanks to their flexibility for transfer learning, and easy tools available on the Web \cite{Caffe2017}.
An artificial Neural Network (NN), composed of artificial neurons \ref{fig:NNoutput}, or nodes, can be used for solving artificial intelligence (AI) problems. NNs are biologically inspired, where a neural network is a network or circuit of neurons in the brain. The connections of the biological neuron are modeled as weights: a positive weight reflects an excitatory connection, while negative values mean inhibitory connections. In 1983 Geoff Hinton, now an emeritus professor at University of Toronto, co-invented Boltzmann machines, \cite{boltzmann} one of the first types of neural networks to use statistical probabilities, then updating the strength of the connections within a neural network with backpropagation. \cite{backpropagation}
In the late-1970s and early-1980s, Hinton began working with neural networks when they were deeply unfashionable, because most computer scientists believed the technique was a dead end, while a better approach to Artificial Intelligence (AI) could be to explicitly encode human expertise in rules sets. Today we know that deep neural networks using backpropagation underpin most advances in AI, from Facebook friends automatic tagging, to the voice recognition capabilities of Amazon Alexa and Google Home, to its translation capability from previously difficult languages, such as Mandarin.
LeCun, then, was a post-doc with Hinton's supervision, developing Convolutional Neural Networks as an improvement of the work on backpropagation. Bengio, who worked with LeCun on computer vision at Bell Labs, applied neural networks to natural language processing, leading to enormous advances in computer translation. Recently, he also built a model to allow neural networks to create novel and realistic images.
In March 2019, Hinton, LeCun and Bengio received together the Turing Award, considered the Nobel prize for computing, for their advances in Artificial Intelligence with Deep Learning. \cite{DeepLearning}
As we can see in Fig. \ref{fig:NNoutput}, a single perceptron (i.e., the NN atom) takes several binary inputs, $x_1, x_2,…,x_n$ and produces a single binary output.
Weights $w_1, w_2,…,w_n$ can be introduced to express the importance of the respective inputs to the output. The neuron's output, 0 or 1, is determined whether the weighted sum $\sum_{j} w_jx_j$ is less/greater than a threshold parameter value of the network:
\begin{equation}
output=
\begin{cases}
 0 & if \sum_{j}^{}w_jx_j \\
 1 & if \sum_{j} w_jx_j
\end{cases}
\end{equation}

By varying the weights and threshold, we can get different models of decision-making, thus different devices device capable to make decisions by weighting up evidence. A real NN will have several perceptrons in each column, and several cascade columns, where each columns is called a \textit{layer}.
A several-layers NN of perceptrons can engage sophisticated decision-making, adding variations to the comparison to the threshold. Several types of layers can be adapted to different calculation aims.

\subsection*{Convolutional Neural Networks}
A Convolutional Neural Network (CNN) is a class of deep neural networks, most commonly applied to analyzing visual content, with excellent results on image recognition, segmentation, detection and retrieval.\cite{AlexNet,farabet2013} The key enabling factors behind such relevant results were principally techniques to scale up the networks to millions of parameters, where labeled data sets are needed to support the learning process. CNNs are able, under such conditions, to learn powerful and interpretative image features.
Convolutional layers apply an operation of convolution to the input, which emulates the response of an individual neuron to visual stimuli, processes data only for its receptive field. A set of kernels (i.e., learning parameters), with a small receptive field, extend through the full depth of the input volume. A forward pass convolutes each filter across width and height of the volume in input, calculating the dot product between the filter entries and the input, thus producing a 2-dimensional activation map. The network results to learn filters activating when some specific type of feature is detected in a particular position of the input.
Although fully connected neural networks can be used to learn features as well as classify data, a relevant amount of neurons is necessary due to the large input sizes of images, where compression is not always a good idea because any pixel may be relevant. E.g., a fully connected layer for an image of size 100 x 100 will have 10000 weights for each neuron. The  operation of convolution offers a great solution to the problem, so tat the network can be deeper with fewer parameters. E.g., tiling regions of size 5 x 5, regardless of image size, having the same shared weights, require just 25 kernels. The problem of exploding gradients in traditional NNs with many layers is solved using backpropagation.
Convolutional networks may also include local or global pooling layers, to reduce data dimension using a combination of the outputs of neuron clusters obtaining one neuron in the following layer.
\begin{figure}
\centering
\includegraphics[scale=0.7]{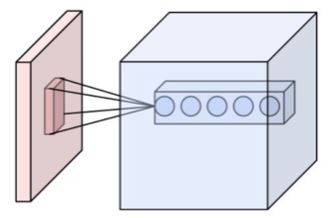}
\caption{A single CNN layer}
\label{fig:CNN}
\end{figure}

\begin{figure}
\centering
\includegraphics[width=\linewidth]{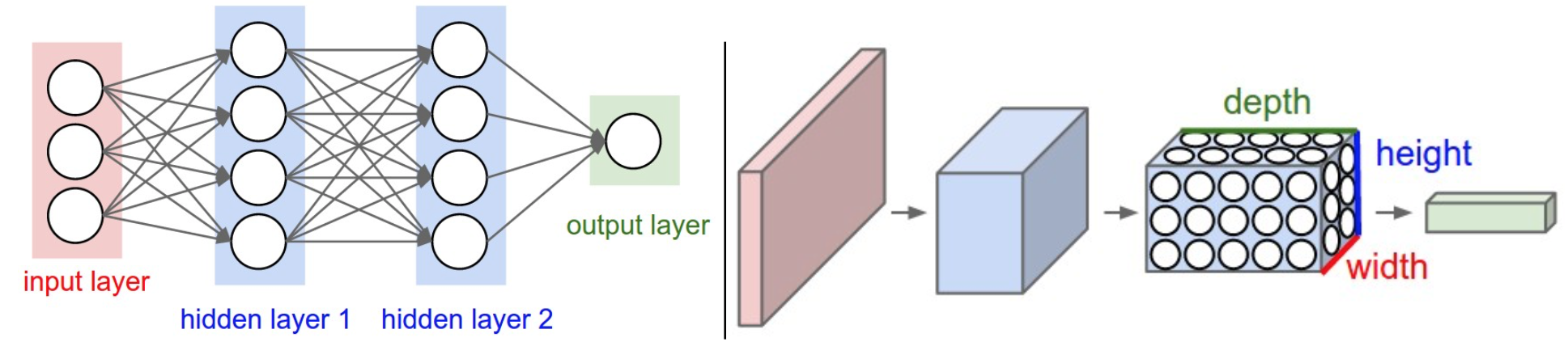}
\caption{Left: A regular 3-layer Neural Network. Right: A Convolutional Network arranges its neurons in three dimensions (width, height, depth). The input layer holds the image.}
\label{fig:threeLayersCNN}
\end{figure}
The neurons in a layer will be connected to a small region of the previous layer, as illustrated in Fig. \ref{fig:CNN}, instead of all of the neurons, as happens in the fully-connected layer, which connects all the neurons in one layer to every neuron in another layer.
A simple CNN is a sequence of layers, each of which transforms one volume of activations to another through a differentiable function. Typically, three types of layers are used: Convolutional Layer, Pooling Layer, and Fully-Connected Layer, which are then stacked together to form a full CNN architecture (see Fig. \ref{fig:threeLayersCNN}). In this way, CNN transforms the original image layer by layer from the original pixel values to the final class scores. Note that some layers contain parameters, and others do not. \cite{mezzetti} In particular, the convolutional and the fully-connected layers perform transformations as a function of both the activations in the input volume, both the the weights and biases of the neurons (i.e. the parameters). On the other hand, the pooling and \textit{RELU} layers, which apply an element-wise activation function, such as the $\max(0,x)$, will implement a fixed function.

\subsection*{Artificial Intelligence assisting Health Care}
For computerized health care assisting, multidisciplinary studies in Artificial Intelligence, Augmented Reality and Robotics stressed out the importance of computer science for automatizing real-life tasks for assistive and learning objects, \cite{nextRevolution} such as detecting words from labial movements (i.e. automated lip detection) \cite{lipdetection}, Virtual reality for prosthetic training \cite{prosthetics} or neural telerehabilitation of patients with stroke \cite{vrrehabilitation}, vocal interfaces for robotics applications \cite{vocalinterfaces}.
As an application of complex networks, it is possible to predict bacteria diffusion patterns, \cite{bacterialdiffusion}\cite{QCN} as well as epidemiology data \cite{epidemiology}, having a viral spread. To be mentioned, huge advances are happening on medical image recognition and multi-stage feature selection for classification of cancer data \cite{cancerdata}, and of text corpora for medical or patient feedback in social networks.
One of the most promising advances of recent years for AI-assisted health care is the opportunity to have light-implementation Mobile Apps, that can be quickly developed to be used in a friendly manner \cite{Gervasi2009}, to assist and support disabled users for communication and learning tasks. Such applications can be run directly on personal smartphones or wearable devices, for health monitoring and prognosis \cite{wearables} as well as for interactive support for people with disabilities or conditions that can influence communication and learning, such as autism spectrum disorders \cite{autism}. Using cloud services or networks in the Internet of Things (IoT), makes possible both to connect such devices to high capability servers, both to collect data in a distributed collaborative perspective \cite{federationscience}, in order to feed big knowledge-bases, and increase the capability of the single object, i.e. of its owner, as a member of a vast interactive collective dynamic knowledge (i.e. a Big Data) network.

\subsection*{Affective Computing and Emotion Recognition}
Multidisciplinary approaches recently stressed out the importance of recognizing and extracting affective and mental states, in particular emotions, for communication, understanding, and supporting humans in any task with automated detectors and artificial assistants having machine emotional intelligence \cite{Picard2001}. In real-life problems, individuals transform overwhelming amounts of heterogeneous data in a manageable and personalized subset of classified items. The process of recognition of moods and sentiments is mostly complex. Recent research underlines that primary emotional states such as happiness, sadness, anger, disgust, or neutral state \cite{mindcybernetics} can be recognized based on text,\cite{Biondi2016} physiological clues such as heart rate, skin conductance and face expression, differently from sentiment, moods and affect, which are more complex states and can be better managed with a multidimensional approach \cite{Vallverdu2017} \cite{Poggioni2013}. Since Rosalind Picard defined the challenges for Affective Computing in 2003 \cite{challenges}, numerous advances have been made in the task of emotion recognition, such as defining collective influence of emotions expressed online \cite{collectiveemotions}, stating that emotional expressiveness is the crucial fuel that sustains communities; studying cultural aspects of emotions in art \cite{emoart} and its variations; create emotionally engaging experiences in games \cite{emogames}, where affective changes are crucial to the conscious experience of the world around us.
Some of the more ethical and critical challenges defined by Rosalind Picard, however, remain open. For example, many of the modalities for emotion recognition (e.g., blood chemistry, brain activity, neurotransmitters) are not readily available, commercial tools are limited \cite{commercialtools}, data sets for training are not general \cite{twitterdata} and people's emotion is so idiosyncratic and variable, that there is difficult to recognize an individual's emotional states from available data \cite{challenges}. Moreover, the challenge to use Affective Computing to help people, e.g., with self-aid tools is not widely faced in research, preferring applications to marketing. \cite{Milani2019}

\section{The proposed Emotion Recognition engine}
We based our Emotion Recognition Engine on popular open-source libraries: the image processing features are provided by OpenCV \cite{OpenCV}, version 3.1.0.
First, the software recognizes the presence of a face in the image; when a face is found, the algorithm looks for the mouth location and extracts the corresponding subframe. \cite{Lewis2016} Both tasks are carried out using Haar Feature-based Classifiers \cite{Haar}, which guarantee fast execution and promising performance.
A face detection pre-trained classifier is integrated into OpenCV; the mouth classifier, instead, is the one used in \cite{mouthdec}. During the training phase, samples of the subject are captured from the camera at regular intervals (or fed from disk) and used to produce a set of mouth snapshots. Such snapshots will form, after shuffling, the training, validation, and test set, with the first two used to train the networks with the Deep Learning framework Caffe \cite{Caffe2017}. The remaining images are subsequently used to test the performances of the networks. The system has been designed to perform both offline and online recognition, i.e. recognize emotions from a series of pre-stored images or directly from a video feed.

\subsection{The structure of the EmEx2 CNN}

In order to have a direct approach to the world of conundrum neural networks, the EmEx \cite{emexIosPress} approach, which we used in part of our tests, focuses on detecting user-centered emotions from the mouth. Network layers are set up to extract the specific information of the image data, accurately setting the parameters to have a valid recognition. In our CNN, training data are labeled with emotions, and the results of the layer computation are evaluated in terms of accuracy and loss.
The neural network architecture is based on the popular \textit{LeNet-5},\cite{lenet}, consisting of several layers connected. The main level is for the \textit{data set} and the corresponding tags. The second layer is the \textit{convoluted layer}, where the convolution operations are performed on the input images, extracting features about each frame in each class. Then, a \textit{pooling layer} is used to reduce the parameter magnitude, reducing in width and height, the volume of previously created data, with a time gain in computation. For scaling, a max function of a variable set is used. Different convolutional and pooling layers follow. An \textit{inner product layer} \textit{innerProduct} groups the information in a single numeric value, to be processed again in the following phases.
The system is now capable of returning a vector representation of neurons, and it will no longer be possible to apply unambiguous layers. Another layer of innerProduct is then applied to put the layers in sequence. Thus, the last one will have an output parameter that equal to the number of classes needed for the classification. The $K$ final values will be the parameters of a probability function that allows the final classification.
In the training phase, the network ends with an \textit{Accuracy layer} for network accuracy calculation, and with a \textit{Loss layer }for the calculation of the error function needed for a correct and useful training phase. In the classification phase, instead, a \textit{SoftMax layer} is the final layer to classify new images, which are not included in the data set (i.e., the test set images in our experiments). This layer calculates the likelihood of the most appropriate class in the grading phase, and therefore, its output represents the final solution.

\subsection{The structure of the AlexNet CNN}\label{AlexNet}
AlexNet \cite{AlexNet}, the second network that we used in our experiments, is a network presented as a winner of the 2012 ImageNet Large-Scale Visual Recognition Challenge (ILSVRC), on the ImageNet \cite{ImageNet} data set, which includes $\approx 1.2$ million pictures representing 1000 different objects in over 22000 categories. \cite{weiliu89ssd} Feed-forward networks could offer the power needed for such a huge data set, requiring much preprocessing work. Using still modern techniques, such as data augmentation and dropout, AlexaNet exploited the benefits of CNN and backed them up with record-breaking performance in the competition.
The AlexaNet CNN is used in several applications; it consists of five convolutional layers, followed by three fully connected layers. Also, three max-pooling layers are inserted after, respectively, the first, second, and fifth convolutional layer, while the first two fully-connected layers are followed by a dropout layer, to avoid overfitting.

\section{Image collection and Training phase}
The first data set is a generalized faces data set, including faces from different ethnicity, gender, age: the \textit{10k US Adult Faces database} \cite{ 10k US Adult Faces database} from the Maryland Laboratory of Brain and Cognition of the USA National Institute of Mental Health, which includes 637 faces images labeled with \textit{Neutral} state and 1511 with \textit{Joy} emotion.\\
For the experiments regarding a single user, we collected images for \textit{Joy}, \textit{Disgust}, and \textit{Neutral}.
During the training phase, the subject was presented with a list of videos and pictures, selected to elicit a particular emotion in the audience. In particular, a set of 62 short videos was used to elicit \textit{Joy}, whereas 140 images were selected for \textit{Disgust}. The participants were asked to sit, one by one, and watch the videos/images, while their reactions were recorded by the camera. Later, samples were extracted from the sections of the videos, which showed an evident reaction to the stimuli, at a rate of three frames per second. For the \textit{Neutral} state, the test subject's expression was recorded watching relaxing images, where no particular emotion was elicited. As a basic rule, we decided that no media could be watched more than once, as the reaction would not be spontaneous anymore in case of multiple views.
The collected samples were then shuffled, to equitably distribute frames belonging to the same sequences between training set, validation set, and test set.

\section{Experiments design and results}
Three experiments were performed for this work, using the previously described two networks, i.e., AlexNet and Emex.
In the first experiment, the networks were trained and tested on the data set composed of samples that we collected from our test subjects; in the second experiment, the same test was repeated on the  \textit{10k US Adult Faces} database. 
Finally, a cross-domain experiment was conducted, testing the networks trained on  10k US Adult Faces database sample to recognize emotions in our single-users data sets.
\subsection{Single-user test}\label{singleTest}
The first test was conducted on the samples collected from each test subjects, in order to see how the networks perform training and testing on the same user in different conditions, e.g., before and after a degenerative pathology, which may prevent the patient from expressing his own emotions and related needs in words.
\begin{table}
\caption{Results of tests using both test networks on the single-user data set}\label{res1}
\begin{tabular}{|l|l|l|l|}
\hline
Network& Training Steps & Accuracy &Micro-Averaged F1\\
\hline
 AlexNet& 50& 0.5437& 0.5437\\
 AlexNet& 100& 0.5340& 0.5340\\
 AlexNet& 150& 1& 1\\
 AlexNet& 200& 0.6484& 0.6484\\
 \hline
 EmEx& 50& 1& 1\\
 EmEx& 100& 1& 1\\
 EmEx& 150& 1& 1\\
 EmEx& 200& 1& 1\\
\hline
\end{tabular}
\end{table}
Results, shown in Table \ref{res1}, show that both networks easily overfit. During the training phase, a perfect accuracy (i.e., 1) was achieved after a few iterations: 50 for the EmEx network, and 150 for the Alexnet network. Further iterations were not necessary, because both networks showed a constant behavior, correctly classifying all the test images. The different training time to obtain the best performances is due to the much higher complexity of the AlexNet network with respect to EmEx, \cite{Riganelli2017}\cite{emexIosPress} in terms of the number of parameters to be optimized. AlexNet was originally designed for a much more complex problem, as stated in section \ref{AlexNet}, i.e., the identification of objects belonging to an extremely high number of classes. It is worth noticing that, although our task was quicker to tackle, inter-class differences may be less evident for our task than for the original ImageNet data. Therefore, our problem is more difficult to solve. Furthermore, the number of training samples used in our experiment was purposefully small, to assess performances in a context where high computing capabilities and data sets are not available, e.g., where the user can train emotional expressions on a mobile environment or a common desktop/laptop, with a relatively small number of images. Moreover, if such images are shot through a video, they will have less intra-dataset differences. 

\subsection{Multiple-users test}
\begin{figure}
\includegraphics[width=\textwidth]{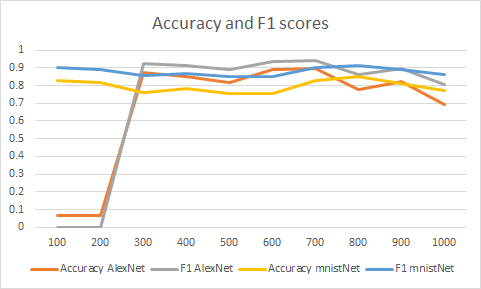}
\caption{Recognition performance of the two networks on the test samples}
\label{fig:graphtest2}
\end{figure}

The second test was performed on the \textit{10k US Adult Faces database} \cite{ 10k US Adult Faces database}, including multiple-users images. This test includes only \textit{Joy} and \textit{Neutral}, due to the lack of enough training samples for \textit{Disgust}.
For both classes, 444 samples were included in the training set, for a total of 888 images, while the validation test set comprised 56 images per class. All the images that were left out, i.e., 814 for joy and 56 for neutral, were used to calculate the metrics. The settings used for the network training, which differ from the original ones, are: 
\begin{itemize}
  \item Train batch size: 10
  \item Test batch size: 16
  \item Test iterations: 7
  \item Test interval: 50
  \item Maximum number of iterations: 1000
  \item Random state seed: 1234
\end{itemize}
The ADAM \cite{adam} optimizer was used on both networks.
\textit{AlexNet} achieved a maximum training accuracy of $\approx0.84$ after 400 iterations, while \textit{EmEx} peaked at a higher $\approx0.91$ score after 800 iterations. However, as shown in the results, AlexNet holds a better generalization ability, thanks to its complexity, achieving both higher accuracy and F1 scores with respect to the EmEx network. Complete results are reported in table \ref{restest2} and figure \ref{fig:graphtest2}.
\begin{table}
\caption{Results of tests using the two networks on the  10k US Adult Faces database images. Best figures for AlexNet in bold; for EmEx in italic bold}
\label{restest2}
\begin{tabular}{|l|l|l|l|l|}
\hline
 &\multicolumn{2}{c|}{AlexNet}&\multicolumn{2}{c|}{EmEx}\\
\hline
Steps& Accuracy & F1& Accuracy & F1\\
\hline
100&0.0644&0.0000&0.8264&0.8993\\
200&0.0644&0.0000&0.8161&0.8922\\
300&0.8713&0.9272&0.7621&0.8549\\
400&0.8506&0.9140&0.7816&0.8684\\
500&0.8172&0.8923&0.7575&0.8517\\
600&0.8897&0.9385&0.7540&0.8491\\
700&\textbf{0.8966}&\textbf{0.9426}&0.8276&0.8992\\
800&0.7770&0.8651&\textit{\textbf{0.8517}}&\textit{\textbf{0.9145}}\\
900&0.8207&0.8949&0.8115&0.8886\\
1000&0.6931&0.8044&0.7701&0.8603\\
\hline
\end{tabular}
\end{table}
\subsection{Cross Test}
Both the AlexNet and EmEx networks trained on the 10k US Adult Faces database were tested on the data set created for the Single-user test in section \ref{singleTest}; results are reported in table \ref{restest3}.
\begin{table}
\caption{Results of tests using the two networks trained on the 10k US Adult Faces database on the Single-user data set. Best figures for AlexNet in bold, for EmEx in italic bold}
\label{restest3}
\begin{tabular}{|l|l|l|l|l|}
\hline
 &\multicolumn{2}{c|}{AlexNet}&\multicolumn{2}{c|}{EmEx}\\
\hline
Steps& Accuracy & F1& Accuracy & F1\\
\hline
100 &0.5054 &0 &0.6344 &0.6793\\
200 &0.7098 &0.5846 &0.4301&0.4647\\
300 &0.6237 &0.3860 &0.8065 &0.7568\\
400 &0.6990 &0.5625 &0.8602 &0.8354\\
500 &0.5484 &0.2500 &\textit{\textbf{0.8925}} &\textit{\textbf{0.8781}}\\
600 &0.6667 &0.4918 &0.8495 &0.8205\\
700 &0.6667 &0.4918 &0.8280 &0.7895\\
800 &0.6129 &0.3571 &0.8172 &0.7733\\
900 &\textbf{0.7204} &\textbf{0.6061} &0.8172 &0.7733\\
1000 &0.5699 &0.2308 &0.8172 &0.7733\\
\hline
\end{tabular}
\end{table}
Interestingly, the EmEx network performed consistingly better than AlexNet, showing a better generalization capability in a completely different environment from the one it was trained for, e.g., with respect to light, user position, and image quality. This is probably due to the ability of AlexNet to better adapt to the peculiar characteristics of the data set it is trained on, thanks to its complexity, but having problems in a fairly different settings when no samples are given.

\section{Conclusions and future work}
In this work, we described a framework for Emotion Recognition from mouth expressions. Experiments, conducted on both single-user and generalized faces data sets, show good recognition performances of the framework, which can correctly identify the chosen emotions, using limited computational resources and doing it both online and offline.
Results show that mouth expressions play an essential role in defining the emotion conveyed by the subject, and can be exploited with low computational power and complexity of systems. Both the tested networks achieved high recognition performances, with the AlexNet network better adapting to the single data set, and a seemingly better ability of the EmEx network to generalize the domain.
Future works will investigate the ability of the framework in recognizing more emotions, and include the publication of our single-users image collection.

\end{document}